\documentclass{article}
\usepackage{bm}
\usepackage{amsmath}
\usepackage{amssymb}
\usepackage{graphicx}
\def\x{\bm{x}}
\newcommand{\ind}{\mathrel{\perp\!\!\!\perp}}
\usepackage[preprint]{neurips_2022}

\usepackage[utf8]{inputenc} % allow utf-8 input
\usepackage[T1]{fontenc}    % use 8-bit T1 fonts
\usepackage{hyperref}       % hyperlinks
\usepackage{url}            % simple URL typesetting
\usepackage{booktabs}       % professional-quality tables
\usepackage{amsfonts}       % blackboard math symbols
\usepackage{nicefrac}       % compact symbols for 1/2, etc.
\usepackage{microtype}      % microtypography
\usepackage{xcolor}         % colors
\usepackage{array}
\usepackage{algorithm,algorithmic}
\newcolumntype{H}{>{\setbox0=\hbox\bgroup}c<{\egroup}@{}}

\usepackage{caption}
\usepackage{subcaption}
\usepackage{wrapfig}

\definecolor{darkpastelgreen}{rgb}{0.01, 0.75, 0.24}

\let\subparagraph\paragraph
\let\subparagraph\paragraph
\usepackage[compact]{titlesec}

\titlespacing{\section}{0pt}{1ex}{1ex}
\titlespacing{\subsection}{0pt}{1ex}{0.5ex}
\titlespacing{\subsubsection}{0pt}{0.5ex}{0.5ex}

\DeclareMathSizes{9}{8.2}{5}{3}
\title{Unified Probabilistic Neural Architecture and Weight Ensembling Improves Model Robustness}

\author{Sumegha Premchandar \\
    Michigan State University\\
    \texttt{premchan@msu.edu} \\
    \And
    Sandeep Madireddy\footnote[1]{Corresponding Author}\\
    Argonne National Laboratory\\
    \texttt{smadireddy@anl.gov} \\
    \AND
    Sanket Jantre\\
    Brookhaven National Laboratory\\
    \texttt{sjantre@bnl.gov} \\
    \And
    Prasanna Balaprakash \\
    Argonne National Laboratory\\
    \texttt{pbalapra@anl.gov} 
}

\begin{document}

\maketitle

\begin{abstract}
Robust machine learning models with accurately calibrated uncertainties are crucial for safety-critical applications. Probabilistic machine learning and especially the Bayesian formalism provide a systematic framework to incorporate robustness through the distributional estimates and reason about uncertainty. Recent works have shown that approximate inference approaches that take the weight space uncertainty of neural networks to generate ensemble prediction are the state-of-the-art. However, architecture choices have mostly been ad hoc, which essentially ignores the epistemic uncertainty from the architecture space. To this end, we propose a {\bf U}nified p{\bf r}obabilistic {\bf a}rchitecture and weight {\bf e}nsembling {\bf N}eural {\bf A}rchitecture {\bf S}earch (\emph{UraeNAS}) that leverages advances in probabilistic neural architecture search and approximate Bayesian inference to generate ensembles form the joint distribution of neural network architectures and weights. The proposed approach showed a significant improvement both with in-distribution ($0.86\%$ in accuracy, $42\%$ in ECE) CIFAR-10 and out-of-distribution ($2.43\%$ in accuracy, $30\%$ in ECE) CIFAR-10-C compared to the baseline deterministic approach.
\end{abstract}

\section{Introduction}
Bayesian neural networks have recently seen a lot of interest due to the potential of these models to provide improved predictions with quantified uncertainty and robustness, which is crucial to designing safe and reliable systems~\cite{hendrycks2021unsolved}, especially for safety-critical applications such as autonomous driving, medicine, and scientific applications such as model-based control of nuclear fusion reactors. Even though modern Bayesian neural networks have great potential for robustness, their inference is challenging due to the presence of millions of parameters and a multi-modal landscape. For this reason, approximate inference techniques such as variational inference (VI) and stochastic gradient Markov chain Monte Carlo are being increasingly adopted. However, VI, which typically makes a unimodal approximation of the multimodal posterior, can be limiting. Recent works in the realm of probabilistic deep learning have shown that ensembles of neural networks~\cite{lakshminarayanan2017simple} have shown superior accuracy and robustness properties over learning single models. This kind of ensembling has been shown to be analogous to sampling models from different modes of multimodal Bayesian posteriors~\cite{wilson2020bayesian,jantre2022-SeBayS} and hence enjoys these superior properties.  

While different techniques for ensembling neural networks have been explored, both in the context of Bayesian and non-Bayesian inference, a key limitation is that ensembles are primarily in the weight space, where the architecture of the neural networks is fixed arbitrarily. For example, techniques such as Monte Carlo dropout~\cite{gal2016dropout}, dropConnect~\cite{wan2013regularization}, Swapout~\cite{singh2016swapout}, SSIG~\cite{jantre2021-SSIG} deactivate certain units/connections during training and testing. They are `implicit", as model ensembling is happening internally in a single model and so are efficient, but the gain in robustness is not significant. On the other hand, ``explicit" ensembling techniques such as Deep Ensembles~\cite{lakshminarayanan2017simple}, BatchEnsemble~\cite{wen2020batchensemble},  MIMO~\cite{MIMO} have shown superior accuracy and robustness gains over single models. Considering just the weight-space uncertainty/ensembles can be a limiting assumption since the architecture choice also contributes to the epistemic (model-form) uncertainty of the prediction. The importance of architecture choice over other considerations in Bayesian neural networks has been highlighted with rightness in~\cite{Izmailov-et-al-2021}.

On the other hand, Neural Architecture Search (NAS) has received tremendous attention recently because of its promise to democratize machine learning and enable the learning of custom, data-specific neural architectures. The most popular approaches in this context are reinforcement learning~\cite{zoph2017neural}, Bayesian optimization~\cite{liu2018progressive}, and evolutionary optimization~\cite{real2019regularized}, but usually incur a large computational overhead. More recently, a differential neural architecture search framework, DARTS~\cite{liu2018darts} was proposed that adopts a continuous relaxation of categorical space to facilitate architecture search through gradient-based optimizers. Distribution-based learning of architecture parameters has recently been explored in DrNAS~\cite{chen2021drnas}, BayesNAS~\cite{zhou2019BayesNAS}, BaLeNAS~\cite{zhang2022balenas} to avoid suboptimal exploration observed with deterministic optimization~\cite{zhang2022balenas} by introducing stochasticity and encouraging exploration. However, these works were tasked with learning a point estimate of the architecture and weights rather than uncertainty quantification, ensembling, or robustness.

In this work, we develop {\bf U}nified p{\bf r}obabilistic {\bf a}rchitecture and weight {\bf e}nsembling {\bf N}eural {\bf A}rchitecture {\bf S}earch (\emph{UraeNAS}) to improve the accuracy and robustness of neural network models. We employ a distribution learning approach to differentiable NAS, which allows us to move beyond ad hoc architecture selection and point estimation of architecture parameters to treat them as random variables and estimate their distributions. This property of distribution learning of architectures, when combined with the Bayesian formulation of neural network weights, allows us to characterize the full epistemic uncertainty arising from the modeling choices of neural networks. With \emph{UraeNAS}, we are able to generate rich samples/ensembles from the joint distribution of the architecture and weight parameters, which provides significant improvement in uncertainty/calibration, accuracy, and robustness in both in-distribution and out-of-distribution scenarios compared to deterministic models and weight ensemble models. 

\section{{\bf U}nified p{\bf r}obabilistic {\bf a}rchitecture and weight {\bf e}nsembling {\bf NAS} } 
\subsection{Distributional formulation of differentiable NAS}
In the differentiable NAS setup, the neural network search space is designed by repeatedly stacking building blocks called cells \cite{zoph2017neural,liu2018darts,chen2021drnas}. The cells can be normal cells or reduction cells. Normal cells maintain the spatial resolution of inputs, and reduction cells halve the spatial resolution, but double the number of channels. Different neural network architectures are generated by changing the basic cell structure. Each cell is represented by a Directed Acyclic Graph with N-ordered nodes and E edges. The feature maps are denoted by $\x^{(j)},\ 0\leq j \leq N-1$ and each edge corresponds to an operation $o^{(i,j)}$. The feature map for each node is given by $\x^{(j)} = \sum_{i<j}o^{(i,j)}(\x^{(i)})$, with $\x^{(0)}$ and $\x^{(1)}$ fixed to be the output from the previous two cells. The final output of each cell is obtained by concatenating the outputs of each intermediate node, that is, $(\x^{(2)}, \x^{(3)} \dots ,\x^{(N-1)})$.

The operation selection problem is inherently discrete in nature. However, continuous relaxation of the discrete space \cite{liu2018darts} leads to continuous architecture mixing weights ($\hat{o}^{(i,j)}(\x) = \sum_{o \in O}\theta_{o}^{(i,j)}o(\x)$)  that can be learned through gradient-based optimization. 
The transformed operation $\hat{o}^{(i,j)}$ is a weighted average of the operations selected from a finite candidate space $O$. The input features are denoted by $\x$ and $\theta^{(i,j)}_{o}$ represents the weight of operation $o$ for the edge $(i,j)$. The operation mixing weights $\bm{\theta}^{(i,j)} = (\theta^{(i,j)}_{1}, \theta^{(i,j)}_{2} \dots \theta^{(i,j)}_{|O|})$ belong to a probability simplex, i.e., $\sum_{o \in O}\theta^{(i,j)}_{o} = 1$. Throughout this paper, we use the terms architecture parameters and operation mixing weights interchangeably.

\paragraph{NAS as Bi-level Optimization:}
With a differentiable architecture search (DAS) formulation, NAS can be posed as a 
a bi-level optimization problem over neural network weights $\bm{w}$ and architecture parameters $\bm{\theta}~$\cite{liu2018darts} in the following manner: 
\begin{equation}
\min_{\bm{\theta}}\hspace{0.3em}\mathcal{L}_{val}(\bm{w}^{*}(\bm{\theta}),\bm{\theta}) \quad \text{s.t.} \quad \bm{w}^{*} \in \underset{\bm{w}}{\arg\min}\hspace{0.3em} \mathcal{L}_{train}(\bm{w},\bm{\theta})
\end{equation} 
However, it was observed in recent works~\cite{chen2021drnas} that optimizing directly over architecture parameters can lead to overfitting due to insufficient exploration of the architecture space. To alleviate this, different DAS strategies were employed \cite{chen2021drnas,xie2018snas, Dong2019gdas}. Among them, the most versatile is the distribution learning approach \cite{chen2021drnas} in which the architecture parameters are sampled from a distribution such as the Dirichlet distribution $\bm{\theta}^{(i,j)} \stackrel{iid}{\sim} \text{Dirichlet}(\bm{\beta}^{(i,j)})$ that can inherently satisfy the simplex constraint on the architecture parameters. The expected loss, in this case, can be written as
\begin{equation}\label{eq:drnas}
\min_{\bm{\beta}}\hspace{0.3em} \mathbb{E}_{q(\bm{\theta}|\bm{\beta})}\mathcal{L}_{val}(\bm{w}^{*},\bm{\theta}) + d(\bm{\beta},\hat{\bm{\beta}}) \quad \text{s.t.} \quad \bm{w}^{*}\in \underset{\bm{w}}{\arg\min}\hspace{0.3em} \mathcal{L}_{train}(\bm{w},\bm{\theta})
\end{equation} 
The regularizer term $d(\bm{\beta},\hat{\bm{\beta}})$ is introduced to achieve a trade-off between exploring the space of architectures and retaining stability in optimization. The parameter $\hat{\bm{\beta}} = (1, 1, \dots 1)$ corresponds to the symmetric Dirichlet. Upon learning the optimal hyperparameters $\bm{\beta}^{*}$ the best architecture can be selected by taking the expected values of the learned Dirichlet distribution. The network weights $\bm{w}$ have typically been treated as deterministic quantities~\cite{chen2021drnas} that have been shown to have a limited view of epistemic uncertainty. However, to achieve our goal of improved uncertainty quantification and model robustness, we propose modeling the joint distribution of the architecture parameters and the neural network weights. We elaborate on our methodology in \ref{subsec:jointdis}.

\subsection{Probabilistic joint architecture and weight distribution learning}\label{subsec:jointdis}
\label{para:prior} We model the weights of the neural network with a standard independent Gaussian distribution, that is, $w_{ijk} \sim N(0,1), w_{i_{1}j_{1}k_{1}} \ind w_{i_{2}j_{2}k_{2}}$ if any of $i_{1} \neq i_{2},j_{1} \neq j_{2}, k_{1} \neq k_{2} $. Here, $i$ indexes the cell to which the weight belongs, $j$ indexes the operation, and $k$ indexes the specific weight, given a cell and operation. Similar to \cite{chen2021drnas} we adopt a Dirichlet process prior for architecture parameters $\bm{\theta}^{(i,j)} \stackrel{iid}{\sim} \text{Dirichlet}(\bm{\beta}^{(i,j)})$. For Bayesian inference, the joint posterior distribution would be given by $P(\bm{\theta},\bm{w}|D) \propto P(D|\bm{\theta},\bm{w})\pi(\bm{\theta},\bm{w})$ where $P(D|\bm{\theta},\bm{w})$ is the likelihood and $\pi(\bm{\theta},\bm{w})$ is the prior, as mentioned above. Markov Chain Monte Carlo (MCMC) methods are a natural choice to sample from this intractable posterior since they produce asymptotically exact posterior samples. However, MCMC methods have computational disadvantages in problems with large, high-dimensional datasets. Variational inference (VI) is a scalable approximate Bayesian inference approach but is generally limited in expressivity when it comes to sampling from complex multimodal posteriors. Izmailov et al. (2021) \cite{Izmailov-et-al-2021} indicate that SGMCMC methods are capable of producing samples that are closer to the true posterior than Mean-field VI. This motivates us to adopt the Cyclical-Stochastic Gradient Langevin Descent (cSGLD) introduced in \cite{Zhang2020Cyclical}, to learn the weights of the neural network $\bm{w}$. Due to the nature of bi-level optimization in \ref{eq:drnas}, we can update the architecture and weight parameters alternatively. As demonstrated by \cite{chen2021drnas} the distribution learning framework in \ref{eq:drnas} together with a $l2$ regularizer for $\beta$ works well in practice. Therefore, we use the iterative optimization procedure described below:
\begin{align}
& \bm{\beta}^{(k)} \leftarrow \bm{\beta}^{(k-1)} - \eta\nabla \mathbb{E}_{q(\bm{\theta}|\bm{\beta}^{(k-1)})}\mathcal{L}_{val}(\bm{w}^{(k-1)},\bm{\theta})\\
& \bm{w}^{(k)} \leftarrow  \bm{w}^{(k-1)} -\alpha_{k}\nabla \mathbb{E}_{q(\bm{\theta}|\bm{\beta}^{(k)})}\mathcal{L}_{train}(\bm{w}^{(k-1)},\bm{\theta}) + \sqrt{2 \alpha_{k}}\epsilon_{k}
\end{align}

The first step above is an update for architecture hyperparameters $\bm{\beta}$ using gradient descent with the objective function chosen to be the log-likelihood of the validation data. The second step is to update the weights of the neural network using c-SGLD. Here, $\epsilon_{k}$ is a standard normal random vector and $\alpha_{k}$ is a cyclical learning rate chosen according to a cosine step size schedule. For more details on how we adjust the step size for c-SGLD across the exploration and sampling phase, see \ref{alg:algo1}. The above alternating parameter update approach has an intuitive similarity to Gibbs sampling in which a single stochastic parameter is updated at a time, conditional on all other parameters.

Unlike existing NAS approaches, \emph{UraeNAS} is capable of generating samples from the joint distribution of the architecture and the weights of the network $(\bm{w},\bm{\theta})_{m}, m = 1, 2, \dots \mathcal{E}$ due to the probabilistic nature of the inference. Predictions are generated by averaging across these ensembles. The algorithm \ref{alg:algo1} in Appendix A details the training and ensembling strategy of \emph{UraeNAS}.

\section{Results and Discussions}
\begin{figure}[!ht]
\centering
  \includegraphics[scale = 0.40]{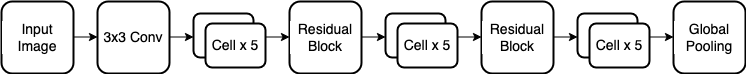}
   \caption{Macro-skeleton for NAS-Bench-201}
  \label{fig:nasbench201}
  \vspace{-4mm}
\end{figure}

We adopted the algorithm-agnostic NAS-Bench-201 search space across all approaches used in our experiments for a fair comparison. NAS-Bench-201 space consists of a macroskeleton formed by stacking normal and reduction cells as shown in Fig.~\ref{fig:nasbench201}. Each cell has 4 nodes, and the operations in the candidate set include zeroize, skip-connect, 1x1 convolution, 3x3 convolution, and 3x3 average pooling. For more details, see \cite{Dong2020NAS-Bench-201}.

In this study, we compare \emph{UraeNAS} with a state-of-the-art probabilistic NAS approach DrNAS that only considers the expected value of the architecture parameters and a network with deterministic weights. We systematically evaluated the effect of architecture and weight ensembling in NAS by evaluating \emph{UraeNAS} in a hierarchy of three strategies. In the first case, we fixed the architecture parameters to the expected value similar to DrNAS and generated ensembles only on the weights (\emph{UraeNAS-w}). In the second case, we generate only architecture ensembles, allowing the weights to be deterministic (\emph{UraeNAS-a}). In the final case, to demonstrate the full potential of \emph{UraeNAS}, we take ensembles from the joint distribution of the architecture and weights. We trained the models on CIFAR-10 data and evaluated their performance in both the in-distribution and out-of-distribution (OoD) cases. The former is done by evaluating the test data, and the latter uses corrupted CIFAR-10, called CIFAR-10-C~\cite{CIFAR-C}, which is the culmination of 20 noise corruptions applied at five different noise intensities. We adopted the three widely used evaluation metrics: Accuracy (Acc), Expected Calibration Error (ECE), and Negative Log-Likelihood (NLL).

\begin{table}[h!]
\small
    \centering
    \vspace{-4mm}
    \caption{Comparison of accuracy and robustness on the CIFAR-10 in-distribution test data and noise corrupted CIFAR-10-C out-of-distribution data.}
    \vspace{1mm}
    \label{tab:accuracy}
    \begin{tabular}{|cHH|c|c|c|c||c|c|c H|}
    \hline
     Approach & Architecture & Weight & Ensembles  & Acc ($\uparrow$)  & ECE ($\downarrow$) & NLL ($\downarrow$) &  cAcc ($\uparrow$) & cECE ($\downarrow$) & cNLL ($\downarrow$) & Dis  \\
     \hline
      \emph{DrNAS} & Prob (point est.) & Deterministic & $1$ & 94.36 & 0.040 & 0.280 & 72.61 & 0.216 & 1.608 & - \\
     \hline
     \emph{UraeNAS-w} & Prob (point est.) & Probabilistic & $10$ & 94.37 & 0.029 & 0.247 & 74.91 & 0.159 & 1.178 & 0.046\\
     \hline
     \emph{UraeNAS-a} & Probabilistic & Deterministic & $10$ & 95.09 & 0.025 & 0.301 & 74.66 & 0.155 & 1.111 & 0.082 \\
     \hline
     \emph{UraeNAS} & Probabilistic & Probabilistic & $10$ & {\color{blue}95.22} & {\color{blue}0.023} & {\color{blue}0.230} & {\color{blue}75.04} & {\color{blue}0.151} & {\color{blue}1.110} & {\color{blue}0.053}\\ 
     \hline
    \end{tabular}
\end{table}

\begin{figure}[h!]
     \centering
     \begin{subfigure}[b]{0.32\textwidth}
         \centering
         \includegraphics[width=\textwidth]{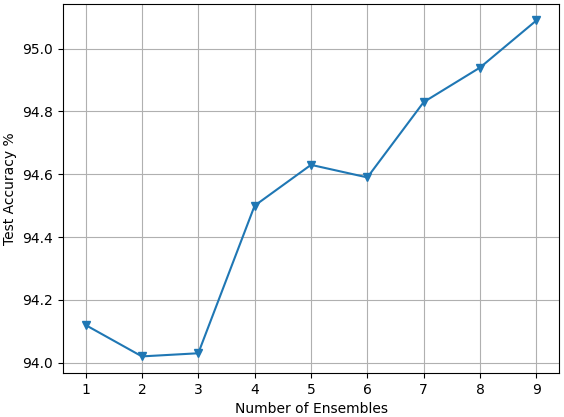}
         \caption{Accuracy}
         \label{fig:acc}
     \end{subfigure}
     \hfill
     \begin{subfigure}[b]{0.32\textwidth}
         \centering
         \includegraphics[width=\textwidth]{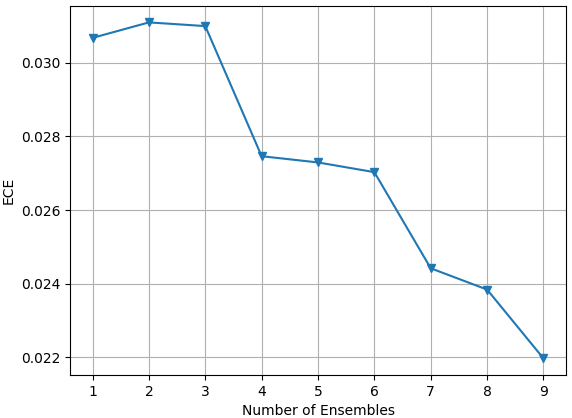}
         \caption{Expected calibration error}
         \label{fig:ece}
     \end{subfigure}
     \hfill
     \begin{subfigure}[b]{0.32\textwidth}
         \centering
         \includegraphics[width=\textwidth]{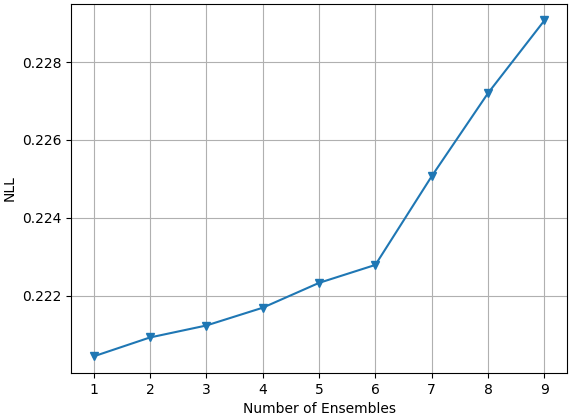}
         \caption{Negative log-likelihood}
         \label{fig:five over x}
     \end{subfigure}
     \vspace{-5mm}
    \caption{Accuracy and calibration evaluation metrics as a function of the number of ensembles.}
    \label{fig:Ensembles}
\end{figure}

The results of these experiments are summarized in Table~\ref{tab:accuracy}, where we see that the weight ensembles improve all three metrics in the in-distribution and OoD cases. When only architecture ensembles are used keeping the weights deterministic, we found further improvement in accuracy and ECE for in-distribution, while in ECE and NLL for the OoD scenario. Lastly, when the ensembles are taken from the joint architecture and weight distribution, we find the highest improvement across the board on all three metrics in the in-distribution and OoD scenarios. The improvement in accuracy is $0.86\%$, ECE is $42\%$, and NLL is $18\%$ for in-distribution, while for the OoD case, the improvement in accuracy is $2.43\%$, in ECE is $30\%$, and NLL is $31\%$ over the baseline approach \emph{DrNAS}. Next, we study the effect of the size of the ensemble on the evaluation metrics for \emph{UraeNAS} and present it in Figure~\ref{fig:Ensembles}, where we find that the metrics improve monotonically (except for a few instances) as the size of the ensemble increases.

\vspace{-7.5mm}
\section{Conclusions and Future Work}
In this work, we proposed the \emph{UraeNAS} approach that leverages probabilistic differentiable neural architecture search and approximate Bayesian inference to learn a joint distribution of the architecture and weight parameters in a neural network. Through our experiments with CIFAR-10 for in-distribution data and corrupted CIFAR-10 for out-of-distribution data, we conclude with the following remarks: (1) weight samples/ensemble improves calibration and accuracy compared to (single) deterministic model;(2) architecture samples improve both accuracy and calibration/robustness over the weight ensemble;(3) joint architecture and weight ensemble improve both accuracy and calibration/robustness over deterministic, weight-only ensembles, and architecture-only ensembles.

As a follow-up work, we will compare \emph{UraeNAS} with other probabilistic weight ensembling approaches~\cite{MIMO} that improve on naive deep ensembles, as well as deterministic architecture ensembling techniques~\cite{egele2021autodeuq} to systematically study the computational cost vs. robustness/accuracy trade-off. Furthermore, we will expand the benchmark data and architecture space to study the generalizability of \emph{UraeNAS}.

\section{Acknowledgments}
This material is based upon work supported by the US
Department of Energy, Office of Science, Office of Fusion
Energy Sciences under award DESC0021203, and Advanced 
Scientific Computing Research through SciDAC Rapids2 
under the contract number DE-AC02-06CH11357.
We gratefully acknowledge the computing resources provided 
on Swing, a high-performance computing cluster operated by 
the Laboratory Computing Resource Center at Argonne National Laboratory.

% \clearpage
% \bibliographystyle{jasa_unsorted}
% \bibliography{refs}

\newpage

\appendix
\section{Appendix}

\begin{algorithm}[h!] 
\caption{{\bf U}nified p{\bf r}obabilistic {\bf a}rchitecture and weight {\bf e}nsembling {\bf NAS} (UraeNAS) } \label{alg:algo1}
\label{alg_ProbNAS}
\small
\begin{algorithmic}[1]
    \STATE {\bfseries Inputs:} training data $\mathcal{D}_t=\{(x_i,y_i)\}_{i=1}^N$, validation data $\mathcal{D}_v=\{(x_i,y_i)\}_{i=1}^N$, operation candidate space $O$, number of architecture ensembles ($M_{\theta}$), number of weight ensembles ($M_w$), model initialization $(\bm{\beta}^{0},\ \bm{w}^{0})$, learning rates $(\eta, \alpha_{0})$, total training epochs $K$, number of cycles $C$, and exploration threshold $r$ for cSGLD.\\
    \STATE {\bfseries Method:} \\
    \textcolor{gray}{\# Train Phase}
    \FOR{$k= 1,2,\dots, K $}
        \STATE Update $\bm{\beta}^{(k)} \leftarrow \bm{\beta}^{(k-1)} - \eta\nabla \mathbb{E}_{q(\bm{\theta}|\bm{\beta}^{(k-1)})}\mathcal{L}_{val}(\bm{w}^{(k-1)},\bm{\theta})$
        \STATE \label{step:sgld} Adjust cyclical learning rate $\alpha_{k} = \frac{\alpha_{0}}{2}\bigg[cos(\frac{\pi mod(k-1,\lceil K/C \rceil)}{\lceil K/C \rceil})+1\bigg]$ 
        % \STATE 
        \IF{ $\frac{mod(k-1,[K/M])}{[K/M]} < r$}
        \STATE \textit{Exploration:} $\bm{w}^{(k)} \leftarrow  \bm{w}^{(k-1)} -\alpha_{k}\nabla \mathbb{E}_{q(\bm{\theta}|\bm{\beta}^{(k)})}\mathcal{L}_{train}(\bm{w}^{(k-1)},\bm{\theta})$
        \ELSE  
        \STATE \textit{Sampling:} $\bm{w}^{(k)} \leftarrow  \bm{w}^{(k-1)} -\alpha_{k}\nabla \mathbb{E}_{q(\bm{\theta}| \bm{\beta}^{(k)})} \mathcal{L}_{train} (\bm{w}^{(k-1)},\bm{\theta}) + \sqrt{2 \alpha_{k}} \epsilon_{k}$, \hspace{3mm} $\epsilon_{k} \sim N(0,I)$ 
        \ENDIF 
        \label{step:endsgld}
    \ENDFOR
    \STATE Store the converged architecture hyper-parameters $\bm{\beta}^{(K)}$.
    \\\textcolor{gray}{\# Evaluation Phase}
    \FOR{$m= 1,2,\dots, M_{\theta}$}
    \STATE Sample $\bm{\theta}_{m} \stackrel{iid}{\sim} \text{Dirichlet}(\bm{\beta}^{(K)}) $ to generate model architecture $m$.
        \FOR{$k= 1,2,\dots, (2K)$}
            \STATE Update weight parameters $\bm{w}^{(k)}$ using steps \ref{step:sgld} to \ref{step:endsgld}.
            \STATE Store last $\lfloor \frac{M_{w}}{C} \rfloor$ weight parameters $\bm{w}^{(k)}$ in \textit{Sampling stage} of each cycle.
        \ENDFOR
    \ENDFOR
    \STATE Load models $\mathcal{M}:(\bm{w},\bm{\theta})_{m}, m = 1,2, \dots \mathcal{E}$,\hspace{5mm} $\mathcal{E} \leq M_{w} \times M_{\theta}$ and generate predictions $y_{m}$. 
    \STATE \textbf{Output:} Stored models corresponding to $(\bm{w}_{m_{1}},\bm{\theta}_{m_{2}})$ \hspace{2mm} $1 \leq m_{1} \leq M_{w}, 1 \leq m_{2} \leq M_{\theta}$, Ensemble predictions $y_{avg} = \frac{1}{m}\sum_{m=1}^{M}y_{m}$
\end{algorithmic}
\end{algorithm}

\end{document}